\UseRawInputEncoding
\documentclass[10pt, conference]{IEEEtran}
\IEEEoverridecommandlockouts
\usepackage{cite}
\usepackage{amsmath,amssymb,amsfonts}
\usepackage{algorithmic}
\usepackage{graphicx}
\usepackage{textcomp}
\usepackage{xcolor}
\usepackage{multirow}

\usepackage[linesnumbered,ruled,vlined]{algorithm2e}

\def\BibTeX{{\rm B\kern-.05em{\sc i\kern-.025em b}\kern-.08em
    T\kern-.1667em\lower.7ex\hbox{E}\kern-.125emX}}

\IEEEpubid{\makebox[\columnwidth]{978-1-6654-3274-0/21/\$31.00 ©2021 IEEE \hfill} \hspace{\columnsep}\makebox[\columnwidth]{ }}
    
\begin{document}

\title{LENS: Layer Distribution Enabled Neural Architecture Search in Edge-Cloud Hierarchies\vspace{-0.4em}}


\author{\IEEEauthorblockN{Mohanad Odema, Nafiul Rashid, Berken Utku Demirel, Mohammad Abdullah Al Faruque}
\IEEEauthorblockA{\textit{Department of Electrical Engineering and Computer Science} \\
\textit{University of California, Irvine, California, USA}\\ 
\textit{\{modema, nafiulr, bdemirel, alfaruqu\}}@uci.edu}\vspace{-8truemm}}

\maketitle
\IEEEpubidadjcol
\begin{abstract}
Edge-Cloud hierarchical systems employing intelligence through Deep Neural Networks (DNNs) endure the dilemma of workload distribution within them. Previous solutions proposed to distribute workloads at runtime according to the state of the surroundings, like the wireless conditions. However, such conditions are usually overlooked at design time. This paper addresses this issue for DNN architectural design by presenting a novel methodology, \textit{LENS}, which administers multi-objective Neural Architecture Search (NAS) for two-tiered systems, where the performance objectives are refashioned to consider the wireless communication parameters. From our experimental search space, we demonstrate that \textit{LENS} improves upon the traditional solution's Pareto set by 76.47\% and 75\% with respect to the energy and latency metrics, respectively. 
\end{abstract}


\section{Introduction and Related Work}
Edge devices are being utilized nowadays in diverse application domains to collect and generate enormous amounts of data. Such applications vary from autonomous vehicles to mobile health, requiring fast and efficient data processing to recognize various events. Usually, Deep Neural Networks (DNNs) are used for processing since they are capable of learning from their previous experiences to make inferences about futuristic occurrences. In that regard, the traditional approach entailed having a DNN running in a centralized cloud, with the data being relayed to it from edge devices. However, the recent trend of laying off some, if not all, of the processing to the edge device has been gaining popularity. One reason for that was to avoid large communication overheads when transferring data over wireless links.

Thus, based on the capabilities of the edge device, the main processing workloads are either distributed between the edge and cloud, or run entirely on a singular endpoint. Furthermore, research works have been addressing how the workloads can adapt dynamically to the varying surrounding conditions. In terms of DNNs, two approaches have been proposed to distribute their workloads in hybrid edge-cloud systems. The first approach in \cite{sieve, torino} proposed to have two separate models on each endpoint, and a lightweight runtime algorithm decides whether to process the data locally or relay it to the cloud depending on the estimated computation costs, network conditions, or input complexity. The other approach in \cite{neurosurgeon} entailed having the same DNN architectural model deployed on both endpoints, and the best partitioning point to split the processing between the edge and the cloud is dynamically adjusted at runtime to adapt to the current state of wireless and server conditions.

On the other hand, Neural Architecture Search (NAS) \cite{NASRL1, NASRL2} has been widely acclaimed as an effective DNN design methodology, in which the DNN design process is automated through conducting a systematic search for the optimal architectural parameters. Each iteration, the candidate architectures are evaluated, and the top-performing ones are retained. Although the main optimization objective in NAS has been to elevate the prediction accuracy, recent works incorporated other optimization objectives besides accuracy like minimizing energy consumption, latency, memory usage, or even hardware design metrics \cite{TEADNN, sparse, HW1, HW2}. This was mainly driven by the high computing demands required of resource-constrained edge devices. Yet thus far, NAS has targeted designing DNNs for deployment on singular platforms, without considering the impact a potential multi-tiered deployment could have.
\vspace{-2.6ex}

\begin{figure}[!htbp]
\begin{center}
{\includegraphics[,width=0.48\textwidth]{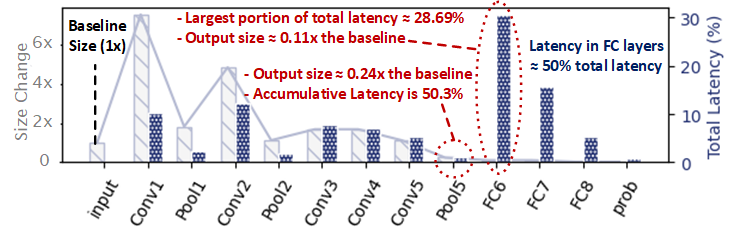}}
\end{center}
\vspace{-3ex}
\caption{Changes in the output feature maps' size and percentage of total latency for each layer in Alexnet}
\label{fig1}
\vspace{-1ex}
\end{figure}

Adapting NAS for multi-tiered hierarchies at design time requires addressing the following research challenges:

\begin{itemize}
    \item How to define the optimal deployment scheme for distributed computation at each tier of the hierarchy? 
    \item How to incorporate the environmental parameters' effect, like expected wireless conditions, into the design process?
\end{itemize}

To address the above-mentioned challenges, a novel NAS-based design methodology for two-tiered hierarchies is proposed in this paper, which employs the following key features:
\begin{itemize}
    \item Performing a per-layer analysis within the optimization context to evaluate architectural candidates based on their best deployment option.
    \item Redefining the performance  objectives to account for the expected wireless communication parameters.
    \item Ensuring that the deployed model of choice can adapt dynamically at runtime to the varying wireless conditions.
\end{itemize}

\section{Motivational Example}
\subsection{Per-layer Analysis of Alexnet}
Figure \ref{fig1} demonstrates how the size of the intermediate feature maps and latency of computation vary for each layer of the Alexnet architecture \cite{alexnet}. For convenience, any activation or normalization layers in this example are fused with their preceding layers as they incur relatively small latency, and the size of feature maps does not change between them. It can be observed from the per-layer analysis that the last 3 Fully Connected (FC) layers take around $50\%$ of the overall execution time. Moreover, the size of the input feature maps to the FC layers is around $4\times$ less than that of the original input. As pointed out in \cite{neurosurgeon}, a potential resource-efficient deployment option for edge-cloud hierarchies can be exploited in which the architecture can be divided between them at the first FC layer's entry input. The efficiency of this solution is dependent on the underlying network conditions and the performance capabilities at both endpoints. Another observation is that until the \textit{Pool5} layer, the feature maps' sizes are larger than the original input size. Therefore, all layers prior to \textit{Pool5} do not represent viable partitioning options because sending the input directly to the cloud would always be a more efficient option.


\subsection{Effect of Wireless Conditions on Alexnet Deployment}

\begin{figure}[!htbp]
\begin{center}
{\includegraphics[,width = 0.48\textwidth]{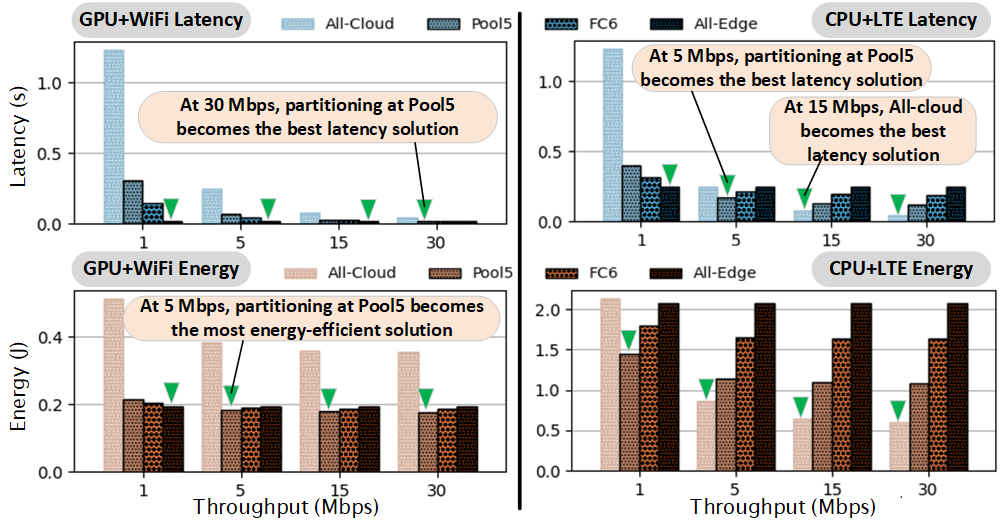}}
\end{center}
\vspace{-4ex}
\caption{The effect of the underlying network conditions on choosing the best partitioning scheme for different device capabilities}
\label{fig2}
\end{figure}

Next, we analyze the total latency and energy consumption for predicting a single input image in Alexnet under different deployment options. Jetson TX2 \cite{jetson} is used as our edge device, and the cloud is assumed of infinite resources. The image is either predicted locally or sent to the cloud at some point to complete its processing. From the previous subsection, \textit{Pool5} and \textit{FC6} represented viable partitioning options for Alexnet alongside the \textit{All-Cloud} and \textit{All-Edge} solutions. Using the Jetson TX2, we evaluate the performance metrics for different deployment options under GPU/WiFi and CPU/LTE combinations at different values of upload throughput $(t_{u})$. As illustrated in Figure \ref{fig2}, it is observed that for each performance metric, the best deployment option varies based on $t_{u}$. For example, to attain the best latency in the GPU/WiFi scenario, the 30 Mbps case prefers partitioning after the \textit{Pool5} layer, contrary to other cases which prefer the \textit{All-Edge} option.

The takeaway from this analysis is that for edge-cloud hierarchies, the same application can have different deployment preferences based on the underlying network conditions. Although previous works address adapting to the network variability at runtime, discrepancies across different regions due to the differences in infrastructure and expected users' bandwidth can have a more significant effect on communication efficiency. Hence, we argue that the expected network conditions should also be considered at design time. Table \ref{table1} shows how the Alexnet would favor different deployment options from our analysis across different regions based on users' average experienced $t_{u}$ values in \cite{opensignal}. 

\section{Preliminaries}
\subsection{Energy and Latency Computation}

For this work, we are considering the scenario in which data can be transmitted from the edge to the cloud. Hence, an application can perform computations locally on the edge or offloads part, if not all, of it to the cloud. Consequentially, we can model the total latency and energy consumption as:
\begin{align}
    L_{total} &= L_{edge} + L_{comm} + L_{cloud} \\
    E_{total} &= E_{edge} + E_{comm} + E_{cloud}
\end{align}
where $L_{edge}$, $E_{edge}$, $L_{cloud}$, and $E_{cloud}$ represent the execution latency and energy consumption at the edge and cloud, respectively. $L_{comm}$ and $E_{comm}$ represent the respective communication latency and energy. As in \cite{neurosurgeon}, \cite{torino}, we care about the energy efficiency from the edge device's perspective, and as the cloud contains much more computation capabilities,  $E_{cloud}$ and $L_{cloud}$ can be neglected with respect to the other factors. For $L_{comm}$ and $E_{comm}$, they are estimated as:
\begin{align}
    L_{comm} &= L_{Tx} + L_{RT} \\
    E_{comm} &= E_{Tx}
\end{align}
where $L_{RT}$ represents the round-trip network latency, while $L_{Tx}$ and $E_{Tx}$ represent the data transmission respective latency and energy consumption which, can be estimated as: 
\begin{align}
    L_{Tx} &= Size(data)/t_{u} \\
    E_{Tx} &= P_{Tx}.L_{Tx} 
\end{align}
in which $Size(data)$, $t_{u}$, and $P_{Tx}$ are the transmission data size, upload throughput, and transmission power, respectively. $P_{Tx}$ can be determined using the power models proposed in \cite{close}, which estimates the power consumption based on the value of $t_{u}$ and the wireless technology used.

\begin{figure*}[!htbp]
\begin{center}
{\includegraphics[,width = 1\textwidth]{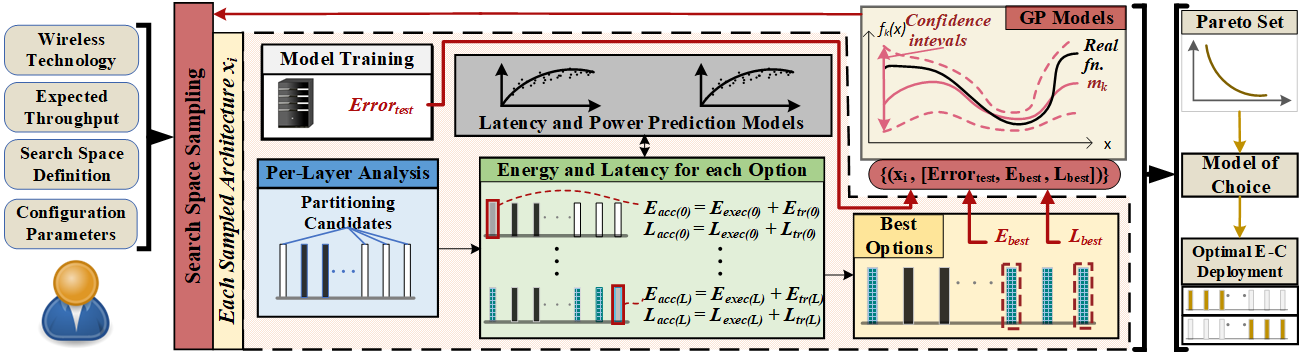}}
\end{center}
\vspace{-2.6ex}
\caption{LENS Design Methodology: Users input the supported wireless technology, the expected wireless conditions, and the search space definition. Sampled candidate architectures have their performance objectives evaluated according to their optimal split, and the evaluations, alongside the test error, are used to update the GP models associated with each objective. The output from \textit{LENS} is a set of Pareto optimal architectures based on their best deployment option.}
\label{fig3}
\vspace{-2ex}
\end{figure*}

\begin{table}[tbp]
		\vspace{-3ex}
		\centering
		\caption{Variability of deployment options across different regions, device capabilities, and performance metrics}
		\label{table1}
		\resizebox{\linewidth}{!}{%
		\begin{tabular}{|c|c|c|c|c|c|} 
			\hline
			\multirow{2}{*}{\textbf{Region}} & \boldmath{$t_{u}$} & \textbf{GPU/WiFi} & \textbf{GPU/WiFi} & \textbf{CPU/LTE} & \textbf{CPU/LTE} \\
			& (Mbps) & (Latency) & (Energy) & (Latency) & (Energy) \\
			\hline
			\textbf{S. Korea} & {16.1} & \textit{{All-Edge}} & \textit{{Pool5}} & \textit{{All-Cloud}} & \textit{{All-Cloud}} \\
			\hline
			\textbf{USA} & {7.5} & \textit{{All-Edge}} & \textit{{Pool5}} & \textit{{Pool5}} & \textit{{All-Cloud}} \\
			\hline
			\textbf{Afghanistan} & {0.7} & \textit{{All-Edge}} & \textit{{All-Edge}} & \textit{{All-Edge}} & \textit{{Pool5}} \\
			\hline
			
			\hline
		\end{tabular}}
		\vspace{-2ex}
\end{table}

\subsection{Multi-Objective Bayesian Optimization}
Bayesian optimization represents an efficient approach to handle black-box optimization problems through a sequential-based exploration. In a multi-objective optimization context, improvement with regard to one objective function \unboldmath{$f_k$} can negatively affect  the others. Thus, the final solution becomes a Pareto optimal set of solutions that dominate all other explored options. A solution \unboldmath{$x^{*}$} is considered Pareto optimal if:
\begin{equation*}
    f_{k}(x^{*}) \leq f_{k}(x) \forall k,x \ \text{and} \ \exists j: f_{j}(x^{*}) < f_{j}(x) \forall x \neq x^{*}
\end{equation*}

In Bayesian optimization, each $f_{k}$ is approximated by a surrogate Gaussian Process (GP) model. Each GP model represents a probability distribution of all possible functions of $f_{k}$ based on previously queried data \unboldmath{$x_{i} \in X_{i=1}^{n}$}. For each $f_{k}$, former evaluations f\unboldmath{$_{kn}=f_{1:n}$} are assumed to be jointly Gaussian with mean ${m_{k}}$ and co-variance $K_{k}$, \textit{i.e.}, f\unboldmath{$_{kn} |$}\unboldmath${x_{1:n}\sim N(m_{k}, K_{k})}$. Through every $m_{k}$ and $K_{k}$, an acquisition function ${\vartheta_{n+1}}$ is constructed to determine the next query point $x_{n+1}$. ${\vartheta_{n+1}}$ is available analytically, making it much cheaper to evaluate than the actual functions, and hence $x_{n+1}$ is determined as: 
\begin{equation}
    x_{n+1} = \arg \max\limits_{x \in X} \ \vartheta_{n+1}
\end{equation}
Once specified, only then are the objective functions evaluated and used to update the GP models through ${m_{k, n+1}}$ and co-variance $K_{k, n+1}$. This loop repeats till the final iteration $N$.

\section{LENS Methodology}
\subsection{Overview}
Figure \ref{fig3} illustrates \textit{LENS} design methodology. Without loss of generality, its NAS framework is based on Multi-Objective Bayesian Optimization (MOBO). Users would be required to specify the search process configuration parameters, in addition to the supported wireless technology type and the expected network conditions. Moreover, the application is specified through its designated search space and dataset. The expected outcome from \textit{LENS} is a Pareto optimal set of architectures minimizing error, latency, and energy consumption.

The search process progresses through sampling a candidate architecture each iteration. The candidate architectural model has its test error estimated after training on the designated dataset. Regarding the performance objectives, the potential layers for partitioning are first determined, then the execution energy and latency values are predicted for each layer and added to those of its preceding layers. Then, the cost of issuing a data transfer to the cloud for each layer is added as well. The minimum values across all the layers' are assigned to their respective objective function evaluations. These evaluations then guide the search process on which areas of the search space it should explore next. Once the final iteration is done, a Pareto optimal set of models is provided for the user.

\begin{figure}
\begin{center}
{\includegraphics[,width = 0.48\textwidth]{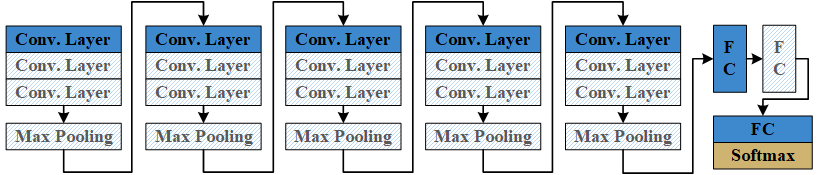}}
\end{center}
\vspace{-2.2ex}
\caption{The defined search space. Lightly-colored layers indicate optional.}
\label{fig4}
\vspace{-2ex}
\end{figure}

\subsection{Search Space}
The benefit of distributing DNN layers between the edge and cloud can only be achieved if the data size at some point within the DNN becomes less than that of the input. Thus, depending on how the search space is defined, a subset of its feasible architectures can encounter enough data shrinkage to make partitioning their optimal deployment option. Whereas for all other architectural candidates, it would be more beneficial to deploy them fully either on the edge or cloud. \textit{LENS} considers the optimal distribution option for that subset of architectures without affecting the other options. Although \textit{LENS} can be adapted to any search space, we demonstrate its merit through an experimental search space derived from the VGG16 \cite{vgg} architecture, which mainly comprises convolutional blocks with stacks of convolutional layers. As shown in Figure \ref{fig4}, our search space comprises 5 convolutional blocks, each is followed by an optional $2\times2$ max-pooling layer. For every block, we vary the number of layers $[1, 2, 3]$, the kernel size $[3, 5, 7]$, and the number of filters $[24, 36, 64, 96, 128, 256]$. After the convolutional blocks, at least one of two Fully Connected (FC) layers can exist with a variable number of neurons for each in $[256, 512, 1024, 2048, 4096, 8192]$. All layers employ \textit{ReLU} as their activation function except for the final FC layer with \textit{softmax} activation. Moreover, batch normalization is applied at all convolutional layers. Lastly, we add the constraint of having at least 4 Pooling layers in each architecture to highlight cases that can benefit from layer distribution. 

\subsection{Layer Performance Prediction Models}
To effectively estimate the execution latency and energy consumption on the target edge device, regression models are constructed to predict the latency and power consumption for each type of layer. We use Caffe \cite{caffe} to construct both the prediction models and the various types of layers. The execution latency can be obtained using Caffe directly while the power consumption is determined using the sensing circuit on the Jetson TX2 board as in \cite{TEADNN}. For each layer's type, different combinations of both layer parameters and input/output feature map sizes are evaluated and used to construct datasets for training the prediction models. Each prediction model would have its input features constructed as in \cite{neurosurgeon}. Once trained, the prediction models can be directly called within \textit{LENS} to estimate the per-layer performance.

\subsection{Algorithm}

Algorithm \ref{alg1} presents the modified performance evaluation functions considering the wireless conditions and the supported wireless technology. After the initialization in \textit{lines 1-3}, \textit{line 4} computes the input and output sizes of the feature maps at each layer based on its architectural parameters. Then, the sizes are used alongside the architectural parameters for the per-layer estimation of latency and power consumption using the regression models as in \textit{lines 5-8}. Furthermore, layers that provide output feature maps whose sizes are less than the entire model's input are selected as candidate partition points in \textit{line 9}. Finally, \textit{lines 10-12} estimate the efficacy of partitioning the model at each candidate through accumulating the on-device execution metrics up to that layer, added to the cost of communicating that layer's output data. The partitioning candidates providing the best performance in regard to each metric are returned. This algorithm is called by Algorithm \ref{alg2} for performance functions evaluations.

\begin{algorithm}[] 
		\footnotesize
		\DontPrintSemicolon
		\caption{Performance Objectives Evaluations}
		\label{alg1}
		\KwIn{Sample: $x$, Wireless Technology: $Tech$, Throughput: $t_{u}$}
		\KwOut{Split Points: $index_{L}, index_{E}$, Latency: $L$, Energy: $E$}
		
		\unboldmath{Initialize $sizes,\ L\_list,\ E\_list,\ L\_acc, E\_acc, \ index\_list$} \\
		\unboldmath{$model = $}Decode($x$)   \tcp*{Construct model} 
		\unboldmath{$\alpha_{u}, \beta_{u} = $}Select\unboldmath{$(Tech)$} \tcp*{Power model Parameters} 
		\unboldmath{$sizes = $}Size\_comp($model$)    \tcp*{Data Sizes per layer}
		\For{\unboldmath{$layer$} \textbf{in} \unboldmath{$model$}}{ 
		    \unboldmath{$l = $}L\_Predict(\unboldmath{$layer, sizes$}) \tcp*{Predict layer latency}
		    \unboldmath{$p = $}P\_Predict(\unboldmath{$layer, sizes$}) \tcp*{Predict layer power}
		    \unboldmath{$L\_list.append(l)$}, \unboldmath{$E\_list.append(p*l)$} \\
		}
		\unboldmath{$index\_list = $}Identify($sizes$) \tcp*{Partition Points}
		\For{\unboldmath{$i$} \textbf{in} \unboldmath{$index\_list$}}{ 
		    \unboldmath{$L\_acc[i] = $}sum(\unboldmath{$L\_list[0:i]$) + }L\_comm\unboldmath{$(sizes[i], t_{u})$} \\
		    \unboldmath{$E\_acc[i] = $}sum(\unboldmath{$E\_list[0:i]$) + }E\_comm\unboldmath{$(sizes[i], t_{u}, \alpha_{u}, \beta_{u})$}
		    }
	    \unboldmath{$index_{L}, L$ = }Minimal($L_{acc}$) \tcp*{Minimal latency}
	    \unboldmath{$index_{E}, E$ = }Minimal($E_{acc}$) \tcp*{Minimal energy}
		
		\textbf{return} \unboldmath{$index_{L}, index_{E}, L, E$ \tcp*{Optimal Values}
		}
\end{algorithm}

Algorithm \ref{alg2} demonstrates the main MOBO-based search for efficiently sampling architectures from the design space. After initialization, \textit{lines 7-11} show that based on functions sampled every iteration from the objective functions' GP posterior models, the acquisition function $\vartheta_{n}$ can be constructed to identify the next query point $x_{n}$. After evaluating the objective functions of $x_{n}$, the Pareto frontier is updated in \textit{lines 12-14}. The computational cost of Algorithm \ref{alg1} is $O(l)$, where $l$ is the number of layers in the architectural model. This cost is minuscule compared to the $O(n^{3})$ cost of a single Bayesian optimization instance \cite{bayes}, where $n$ is the number of samples. \textit{LENS} needs to run only once at design time. 


%

\begin{algorithm}[t] 
		\footnotesize
		\DontPrintSemicolon
		\caption{\textit{LENS} MOBO-based NAS}
		\label{alg2}
		\KwIn{Configuration Variables: \unboldmath{$\{C_{init}, N_{iter}$\}}, Search Space: \unboldmath{$X$}, Objective Functions \unboldmath{$F$}}
		\KwIn{Wireless Technology:  \unboldmath{$Tech$}, Expected Throughput: \unboldmath{$t_{u}$}}
		\KwOut{Pareto Optimal Set: \unboldmath{$X^{*}$}	} 
		\unboldmath{$D = \phi, X^{*} = \phi$} \\
		\tcp{Random initialization starts}
		\For{\unboldmath{$i = 1$} \textbf{to} \unboldmath{$C_{init}$}}{ 
		    \unboldmath{$x_{i} = $} Random(\unboldmath{$X$}) \tcp*{Random sampling}
		    \unboldmath{$Y_{i} = $}Evaluate{$(x_{i}, F, Tech, t_{u})$} \tcp*{Sample Evaluation}
		        
		    \unboldmath{$D = D \ \cup \ (x_{i}, Y_{i})$} \tcp*{Update queried points}
		}
		\unboldmath{$X^{*} = $} Pareto\_init(\unboldmath{$D$}) \tcp*{Initial Pareto frontier}
		\tcp{MOBO starts}
		\For{\unboldmath{$n = 1$} \textbf{to} \unboldmath{$N_{iter}$}} {

		    \For{\unboldmath{$k$} \textbf{in} \unboldmath{$range(F)$}} { 
		        \unboldmath{$f_{k} = $}GP$_{k}$(\unboldmath{$D$}) \tcp*{Sample fn. from posterior}

		        }
		   \unboldmath{$\vartheta_{n} = $}Build\_acq\unboldmath{$(f_{1},..,f_{K}$}) 		   \tcp*{Construct acq. fn.}
		   \unboldmath{$x_{n} = \arg \max\limits_{x \in X} \ \vartheta_{n}$} \tcp*{Maximizer is next query}
		   
		   \unboldmath{$Y_{n} = $}Evaluate{$(x_{n}, F, Tech, t_{u})$} \tcp*{Sample Evaluation} 
		    
		    \unboldmath{$D = D \ \cup \ (x_{n}, Y_{n})$} \tcp*{Update queried points}
		    \unboldmath{$X^{*} = $} Pareto\_update(\unboldmath{$x_{n}, Y_{n}, X^{*}$}) \tcp*{Update Pareto}
		}
		\textbf{return} \unboldmath{$X^{*}$ \tcp*{Final Pareto set}
		}
\end{algorithm}

\begin{figure}[]
\begin{center}
{\includegraphics[,width = 0.48\textwidth]{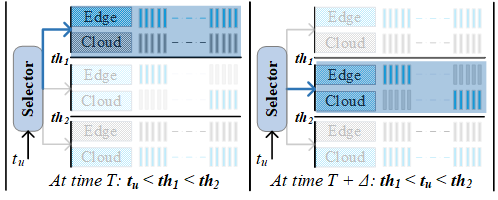}}
\end{center}
\vspace{-3.7ex}
\caption{Runtime system selecting between Partitioned, All-Edge, and All-Cloud. The grayed-out layers indicate inactivity in that mode of operation.}
\label{fig5}
\vspace{-2ex}
\end{figure}

\subsection{Accounting for Runtime Optimizations}


Although \textit{LENS} mainly targets design-time optimization, we also ensure the deployed models remain resilient against any varying network conditions at runtime. Thus before deployment, the model of choice is analyzed to determine the range of $t_{u}$ values over which its best deployment option would remain dominant over other options. To achieve this, each deployment option is compared in a pairwise manner to its counterparts, and the intersection of $t_{u}$ ranges over which it dominates all other options is determined and associated with it. For example, assuming optimizing for latency, the $t_{u}$ threshold after which one deployment option can be dominated by the other can be determined through equating their respective accumulative latency equations in \textit{line 11} of Algorithm {\ref{alg1}}. This is repeated for each pair of options considered for deployment. Once the thresholds are determined, an online throughput tracker can be exploited on the edge device to switch between different deployment options based on the $t_{u}$ value in real-time $O(1)$, as illustrated in Figure \ref{fig5}. 

\section{Experiments}
\textit{LENS}'s multi-objective NAS is built on top of Dragonfly \cite{dragonfly}. Our methodology has no direct competitor, as previous solutions targeting edge-cloud hierarchies assume the DNN architecture is known beforehand, added to the fact that existing NAS works address DNN automated design for single-tiered platforms. Hence, we'll compare \textit{LENS}'s search process against a trivial \textit{Traditional} solution of performing platform-aware NAS for the target edge device, and then applying the optimal distribution of layers between the edge and cloud for its optimal set of architectures. Each Bayesian search experiment is run for 300 iterations on a desktop machine, the average $T_{RT}$ is determined from the average of multiple ping requests to a server machine, and the expected $t_{u}$ is set to $3\ Mbps$. Our test edge device is taken to be the Jetson TX2 with GPU/WiFi support, and the performance prediction models are built accordingly.  For the same architectural design, since only the performance metrics are affected by the type of deployment, the input image size for the performance objectives is set to $224\times224\times3$ (147 kB) to reflect realistic scenarios. For the accuracy objective, we have chosen the CIFAR-10 image dataset to have a relatively fast training process for our models. The CIFAR-10 dataset contains 50,000 training images and 10,000 test images. Out of the training images, we took 5,000 images to be used as the validation set during training. Moderate data augmentation is applied, and each sampled architectural model is trained for 10 epochs. The accuracy objective is evaluated for each candidate model on the test set. 

\begin{figure}[]
\begin{center}
{\includegraphics[,width = 0.48\textwidth]{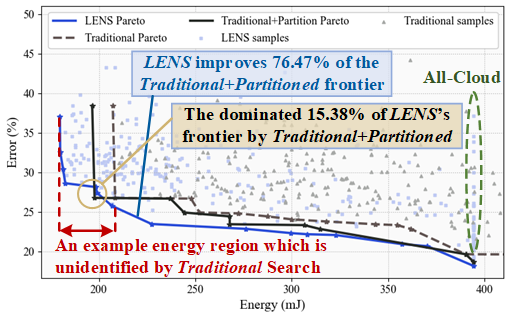}}
\end{center}
\vspace{-3.2ex}
\caption{The Pareto frontiers formed by \textit{LENS}, the \textit{Traditional} solution, and after considering partitioning for the \textit{Traditional}'s frontier.}
\label{fig7}
\vspace{-3.5ex}
\end{figure}

\subsection{LENS vs Traditional}
Figure \ref{fig7} illustrates the explored architectures and the rendered Pareto sets from \textit{LENS} and the \textit{Traditional} solution over the energy and error evaluations. The first observation is that, unlike \textit{LENS}, the \textit{Traditional} solution is unaware of any potential performance gains due to partitioning. For example, no architecture with energy consumption below 207 mJ is identified. As shown, \textit{LENS}'s Pareto frontier dominates completely the \textit{Traditional}'s frontier. After partitioning models in the \textit{Traditional}'s Pareto set, \textit{LENS} still dominates 60\% of the new \textit{Traditional}'s frontier, whereas the new \textit{Traditional}'s frontier dominates 15.38\% of \textit{LENS}'s frontier. Additionally, a combined frontier made from both sets would constitute 76.47\% candidates from \textit{LENS}'s optimal set. The same analysis is applicable for the latency and error trade-off, where \textit{LENS}'s frontier dominates 66.67\% of the partitioned \textit{Traditional} frontier, opposed to 14.28\% of LENS's frontier being dominated. A combined frontier from both is 75\% formed by \textit{LENS}'s models, indicating how \textit{LENS} recognizes the partitioning gains as the search progresses.

\subsection{Partitioning within or after Optimization}

\begin{figure}[]
\begin{center}
{\includegraphics[,width = 0.48\textwidth]{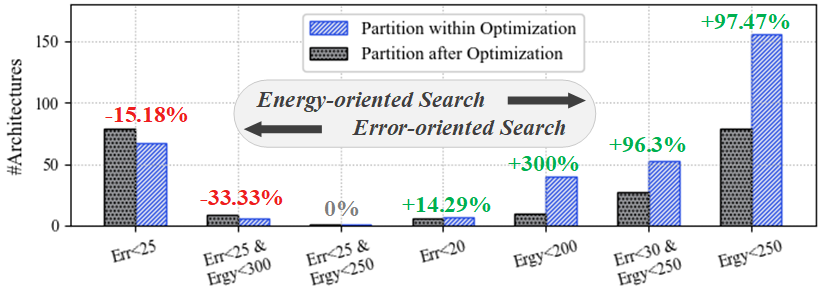}}
\end{center}
\vspace{-3.2ex}
\caption{Number of architectures satisfying the respective conditions. Partitioning within the optimization identifies new opportunities for energy efficiency}
\label{fig6}
\vspace{-3ex}
\end{figure}

We also compare the effectiveness of including the partitioning aspect within the optimization equations themselves, rather than partitioning all the explored solutions after the optimization. This is demonstrated through a comparison of their respective number of architectures that satisfy various criteria of accuracy and energy. Figure \ref{fig6} shows that accounting for partitioning within the optimization equation makes the search more oriented towards improving energy efficiency, demonstrated by the percentage increase in architectures explored that satisfy the various energy criteria, as shown in $Ergy<200$ and $Ergy<250$ conditions. This is because the search acknowledges partitioning improvement potentials, as gains in terms of energy were $10\times$ more than those in terms of accuracy. This has led the search to spend more time exploring further energy gain opportunities before rushing to fine-tune its search with respect to the accuracy objective. Still, the percentage change in the number of architectures satisfying the hardest criterion of $Err<25\ \&\ Ergy<250$ did not change, and even improved for the accuracy condition $Err<20$. This reinforces the claim that partitioning within the optimization provides a more descriptive outlook of the search space.

\subsection{Runtime Analysis}

\begin{figure}[]
\begin{center}
{\includegraphics[,width = 0.48\textwidth]{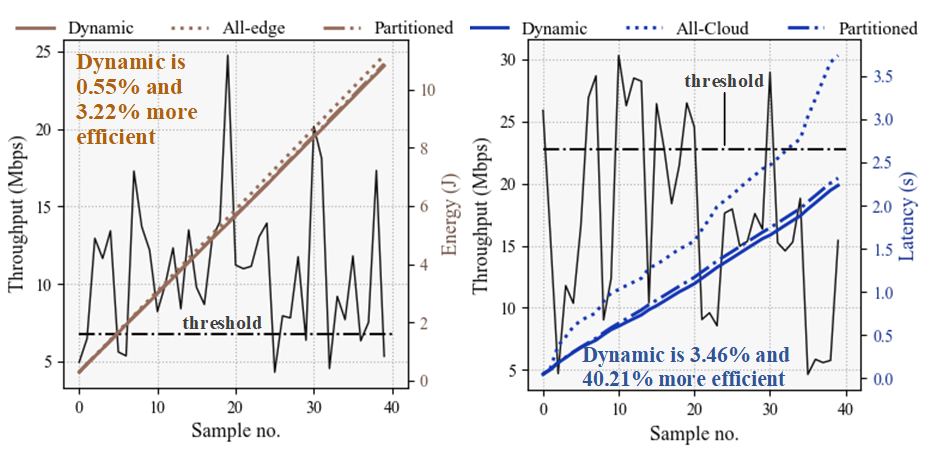}}
\end{center}
\vspace{-3.2ex}
\caption{Changes in accumulative energies and latencies over collected traces of LTE $t_{u}$ for Model $A$ (left) and Model $B$ (right).}
\label{fig8}
\vspace{-2ex}
\end{figure}

To assess the resilience of our models against network variability at runtime, two models, \textit{A} and \textit{B}, are selected from \textit{LENS}'s Pareto frontier for this analysis. We consider each model's best partitioning option, added to the \textit{All-Edge} for model \textit{A}, and the \textit{All-Cloud} for model \textit{B}. The throughput intervals over which each option becomes dominant are identified through pairwise comparison and determining the throughput thresholds. We collect multiple traces of throughput using TestMyNet \cite{testmynet} on a mobile phone for an LTE connection, where $t_{u}$ is measured every 5 minutes for 40 samples. From our analysis, we found that model \textit{A} favors the \textit{partitioned} over \textit{All-edge} regarding energy efficiency whenever $t_{u}>6.77 Mbps$, whereas the latency threshold after which cloud operation is favored for model \textit{B} is defined by $t_{u}>22.77 Mbps$. As illustrated in Figure \ref{fig8}, fixed deployment options for each model are compared against having the throughput tracker dynamically switching between the options at runtime. For model \textit{A}, we find that the \textit{dynamic} option leads to $0.55\%$ and $3.22\%$ increase in the energy efficiency over the the \textit{partitioned} and \textit{All-Edge}, respectively. Speedups were also attained for Model \textit{B} reaching $3.46\%$ and $40.21\%$ over the respective \textit{partitioned} and \textit{All-Cloud} options. The discrepancies in performance gains for each model's case support our claim about the importance of considering the wireless conditions at design time because even though runtime optimization adds to the performance gains, most of the efficiency is already achieved if a model is deployed according to its best option. 

\subsection{Comparison against Other Works}
We also compare \textit{LENS} against other proposed methodologies and techniques that target DNN performance optimization in Edge-Cloud hierarchies. \textit{LENS} is the first work to consider expected wireless conditions at design time. Although, \textit{LENS} and \textit{SIEVE} \cite{sieve} both employ automated design optimizations, they operate within different contexts. First, \textit{LENS} assumes the same architecture partitioned between the edge and cloud while \textit{SIEVE} has two separate architectures on each endpoint. Additionally, \textit{LENS} employs NAS to search through the best deployment options of various architectures to identify the best candidates, whereas \textit{SIEVE} trades off accuracy and efficiency to identify the best representation and the best compression technique for the parameters of each layer in the edge architecture, and then add the necessary hardware to support it. The other works mainly target DNN runtime optimizations for various types of DNNs.


\section{Conclusion}

In this paper, we have introduced \textit{LENS}, which is the first multi-objective NAS design methodology which addresses DNNs' architectural design for Edge-Cloud hierarchies. \textit{LENS} considers the expected wireless conditions at design time within its performance optimization objectives, which allows it to evaluate architectural candidates based on their best layer-distribution scheme. From our experimental search space, we have shown that architectures from \textit{LENS}'s Pareto frontier dominate 60\% of the traditional solution's Pareto set, whereas their combined Pareto set is 76.47\% formed by \textit{LENS}.

\begin{table}[htbp]
		\centering
		\caption{Comparison against other works in terms of the features supported for DNN optimization in Edge-Cloud Hierarchies}
		\label{table2}
		\begin{tabular}{|c|c|c|c|c|} 
			\hline
			\textbf{Supported Features} & \textbf{LENS} & \textbf{NS \cite{neurosurgeon}} & \textbf{SIEVE \cite{sieve}} & \textbf{RNN \cite{torino}} \\
			\hline\hline
			{Design Automation} & {\checkmark} & {-} & {\checkmark} & {-} \\
			\hline
			{NAS support} & {\checkmark} & {-} & {-} & {-} \\
			\hline
			{Wireless expectancy} & \multirow{2}{*}{\checkmark} & \multirow{2}{*}{-} & \multirow{2}{*}{-} & \multirow{2}{*}{-} \\
			{at Design Time} & {} & {} & {} & {} \\
			\hline
			{Multi-Objective} & \multirow{2}{*}{\checkmark} & \multirow{2}{*}{-} & \multirow{2}{*}{\checkmark} & \multirow{2}{*}{-} \\
			{Optimization} & {} & {} & {} & {} \\
			\hline
			{Runtime Optimization} & {\checkmark} & {\checkmark} & {\checkmark} & {\checkmark} \\
			\hline
			{E-C Layer-Partitioning} & {\checkmark} & {\checkmark} & {-} & {-} \\
			\hline
			{Compression} & {-} & {-} & {\checkmark} & {-} \\
			\hline
			{Hardware Optimization} & {-} & {-} & {\checkmark} & {-} \\
			\hline

			\hline
			
		\end{tabular}
		\vspace{-2ex}
\end{table}

\bibliographystyle{IEEEtran}
\bibliography{sample-base}

\end{document}